\documentclass[lettersize,journal]{IEEEtran}
\usepackage{amsmath,amsfonts}
\usepackage{algorithmic}
\usepackage{algorithm}
\usepackage{array}
\usepackage{textcomp}
\usepackage{stfloats}
\usepackage{url}
\usepackage{verbatim}
\usepackage{graphicx}
\usepackage{cite}
\usepackage{booktabs}
\usepackage{multirow}
\usepackage[normalem]{ulem}
\useunder{\uline}{\ul}{}
\usepackage[table]{xcolor}
\usepackage{makecell}
\usepackage{subfigure}

\usepackage[hidelinks]{hyperref}

\newif\ifjournal 

\journalfalse

\begin{document}

\title{OmniFD: A Unified Model for Versatile Face Forgery Detection}

\ifjournal
\else
\author{Haotian Liu, Haoyu Chen, Chenhui Pan, You Hu, Guoying Zhao, and Xiaobai Li}
\fi

\markboth{Journal of \LaTeX\ Class Files,~Vol.~14, No.~8, August~2021}%
{Shell \MakeLowercase{\textit{et al.1}}: A Sample Article Using IEEEtran.cls for IEEE Journals}

\IEEEpubid{0000--0000/00\$00.00~\copyright~2021 IEEE}

\maketitle


\begin{abstract}
Face forgery detection encompasses multiple critical tasks, including identifying forged images and videos and localizing manipulated regions and temporal segments. Current approaches typically employ task-specific models with independent architectures, leading to computational redundancy and ignoring potential correlations across related tasks.
We introduce OmniFD, a unified framework that jointly addresses four core face forgery detection tasks within a single model, i.e., image and video classification, spatial localization, and temporal localization.
Our architecture consists of three principal components: (1) a shared Swin Transformer encoder that extracts unified 4D spatiotemporal representations from both images and video inputs, (2) a cross-task interaction module with learnable queries that dynamically captures inter-task dependencies through attention-based reasoning, and (3) lightweight decoding heads that transform refined representations into corresponding predictions for all FFD tasks.
Extensive experiments demonstrate OmniFD's advantage over task-specific models.
Its unified design leverages multi-task learning to capture generalized representations across tasks, especially enabling fine-grained knowledge transfer that facilitates other tasks. For example, video classification accuracy improves by 4.63\% when image data are incorporated.
Furthermore, by unifying images, videos and the four tasks within one framework, OmniFD achieves superior performance across diverse benchmarks with high efficiency and scalability, e.g., reducing 63\% model parameters and 50\% training time. It establishes a practical and generalizable solution for comprehensive face forgery detection in real-world applications.
The source code is made available at \url{https://github.com/haotianll/OmniFD}.
\end{abstract}


\section{Introduction}
\label{section:introduction}

\begin{figure}[!t]
    \centering
    \includegraphics[width=\linewidth]{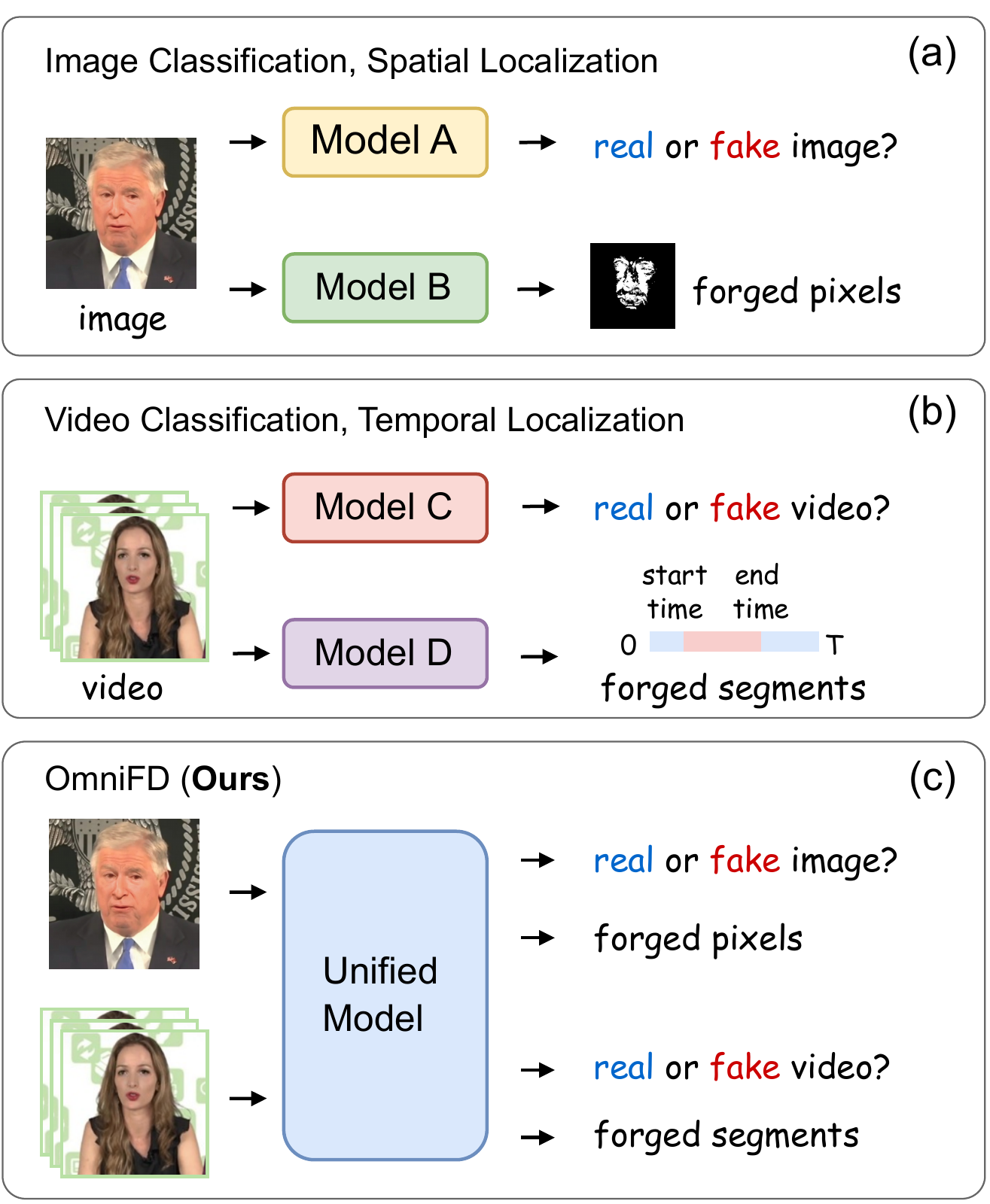}
    \caption{
    Illustration of face forgery detection solutions: (a) and (b) task-specific models for the four tasks, i.e., image classification, spatial localization, video classification and temporal localization; (c) we propose a unified model addressing all the four tasks.
    Our method achieves both superior performance and computational efficiency compared to existing approaches.
}
\label{figure:motivation}
\end{figure}

Face forgery detection (FFD) \cite{faceforensics++_rossler2019, survey_ijcv_juefei2022} identifies whether a face is authentic or fabricated. With the rapid advancements in face synthesis and manipulation technologies \cite{faceforensics++_rossler2019, faceshifter_li2020advancing}, including AI-generated content (AIGC) \cite{GAN_goodfellow2020, stable_diffusion_rombach2022}, FFD has become essential for safeguarding against misinformation, identity fraud, and other malicious applications.

FFD comprises several core tasks, including \textit{Image Forgery Classification} \cite{faceforensics++_rossler2019,efficientnet_tan2019,xception_chollet2017}, \textit{Video Forgery Classification} \cite{forgerynet_he2021, wilddeepfake_zi2020, slowfast_feichtenhofer2019,videoswin_liu2022}, \textit{Spatial Forgery Localization} \cite{dffd_dang2020detection,fmld_kong2022,hrnet_wang2020}, and \textit{Temporal Forgery Localization} \cite{lavdf_cai2023,avdeepfake1m_cai2024,bsn_lin2018,actionformer_zhang2022}. Each task focuses on a distinct aspect of forgery analysis. \textbf{\textit{Image Forgery Classification}} \cite{mesonet_afchar2018, gramnet_liu2020, madd_zhao2021, frequency_qian2020, frequency_li2021, f2trans_miao2023, face_x-ray_li2020, ict_consistency_dong2022, self_blended_shiohara2022, recce_cao2022, m2tr_wang2022, ucf_yan2023, ual_wu2023generalizing, sfdg_wang2023dynamic, laanet_nguyen2024} determines whether a given face image is real or fake, which is the most fundamental FFD task.
\textbf{\textit{Video Forgery Classification}} \cite{ictu_li2018,lips_haliassos2021, biological_ciftci2020, spatiotemporal_gu2021, uniformer_li2022, exploiting_wang2023, tall_xu2023, mintime_coccomini2024, dfgaze_peng2024, dfd_fcg_han2025} processes facial video clips to assess their authenticity, which captures temporal dependencies in video sequences to mitigate the limitations of image-based FFD analysis.
Instead of providing a simple binary classification, \textbf{\textit{Spatial Forgery Localization}} \cite{fmld_kong2022, lavdf_cai2023, dadf_lai2023,upernet_xiao2018, iml_vit_ma2023,mesorch_zhu2025,sparsevit_su2025} identifies manipulated pixels in an image, which aims to enhance practical application and improve interpretability, assisting human users in understanding and verifying forgeries. 
\IEEEpubidadjcol
In addition, \textbf{\textit{Temporal Forgery Localization}} \cite{tadtr_liu2022, tridet_shi2023, re2tal_zhao2023, adatad_liu2024, ba-tfd_cai2022, ummaformer_zhang2023} focuses on detecting manipulated segments within a video clip, where only certain frames are altered. This task addresses the challenge of advanced forgeries in long video clips.
Meanwhile, various FFD benchmarks \cite{faceforensics++_rossler2019, forgerynet_he2021, celeb_li2020celeb, deeperforensics_jiang2020,  wilddeepfake_zi2020, dfdc_dolhansky2020, faceshifter_li2020advancing, deepfakebench_yan2023, fmld_kong2022, lavdf_cai2023, avdeepfake1m_cai2024} have been established for these different tasks to evaluate the performance of FFD methods.
Despite the swift progress, most existing methods are tailored to one specific task.
We summarize the limitations of existing studies as follows:

1) \textit{Redundant and inflexible model designs.} Existing frameworks treat FFD tasks independently, training task-specific models with distinct architectures for each task \cite{faceforensics++_rossler2019, forgerynet_he2021, deepfakebench_yan2023}, as described in Figure \ref{figure:motivation}. This fragmented approach leads to redundancy in both model parameters and training costs, especially when multiple tasks are involved. Moreover, modules specialized for one task are difficult to adapt to others, which limits overall system efficiency.

2) \textit{Potential correlations and benefits among different tasks are overlooked.} For instance, detecting a forged frame in a video can aid in verifying the authenticity of the entire video clip. Although some methods \cite{m2tr_wang2022, forgerynet_he2021, uvif_liu2024} migrate this issue by using additional annotations from related tasks for joint training, they prioritize one primary task and treat others as auxiliary objectives. As a result, the potential of cross-task correlations remains underexploited.

3) \textit{Deployment challenges and poor scalability.} Using multiple specialized models with different architectures \cite{forgerynet_he2021} for FFD tasks lead to inefficiency in deployment, particularly in resource-constrained environments. This limits their applicability in real-world scenarios that require comprehensive FFD analysis, such as the concurrent detection of forgeries and localization of manipulated regions or segments.

\setlength{\subfigcapskip}{10pt}
\setlength{\subfigcapmargin}{0pt}

\begin{figure*}[t]
    \centering
    \subfigure[Performance analysis \label{figure:radar}]
    {
        \begin{minipage}[c]{0.4\linewidth}
            \centering
            \includegraphics[width=\linewidth]{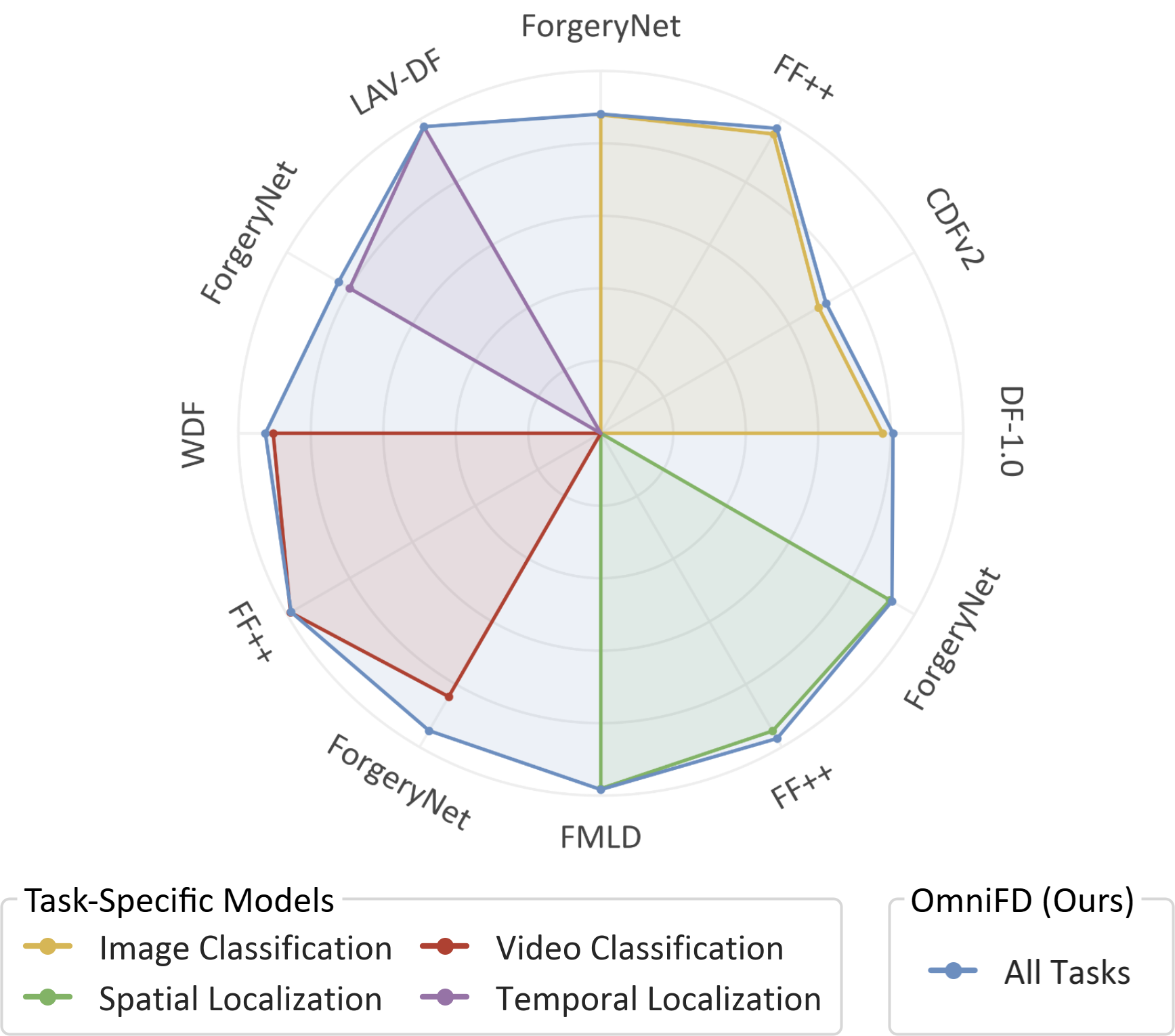}
        \end{minipage}}%
    \hfill
    \subfigure[Efficiency analysis\label{figure:efficiency}]{
        \begin{minipage}[c]{0.58\linewidth}
            \centering
            \includegraphics[width=\linewidth]{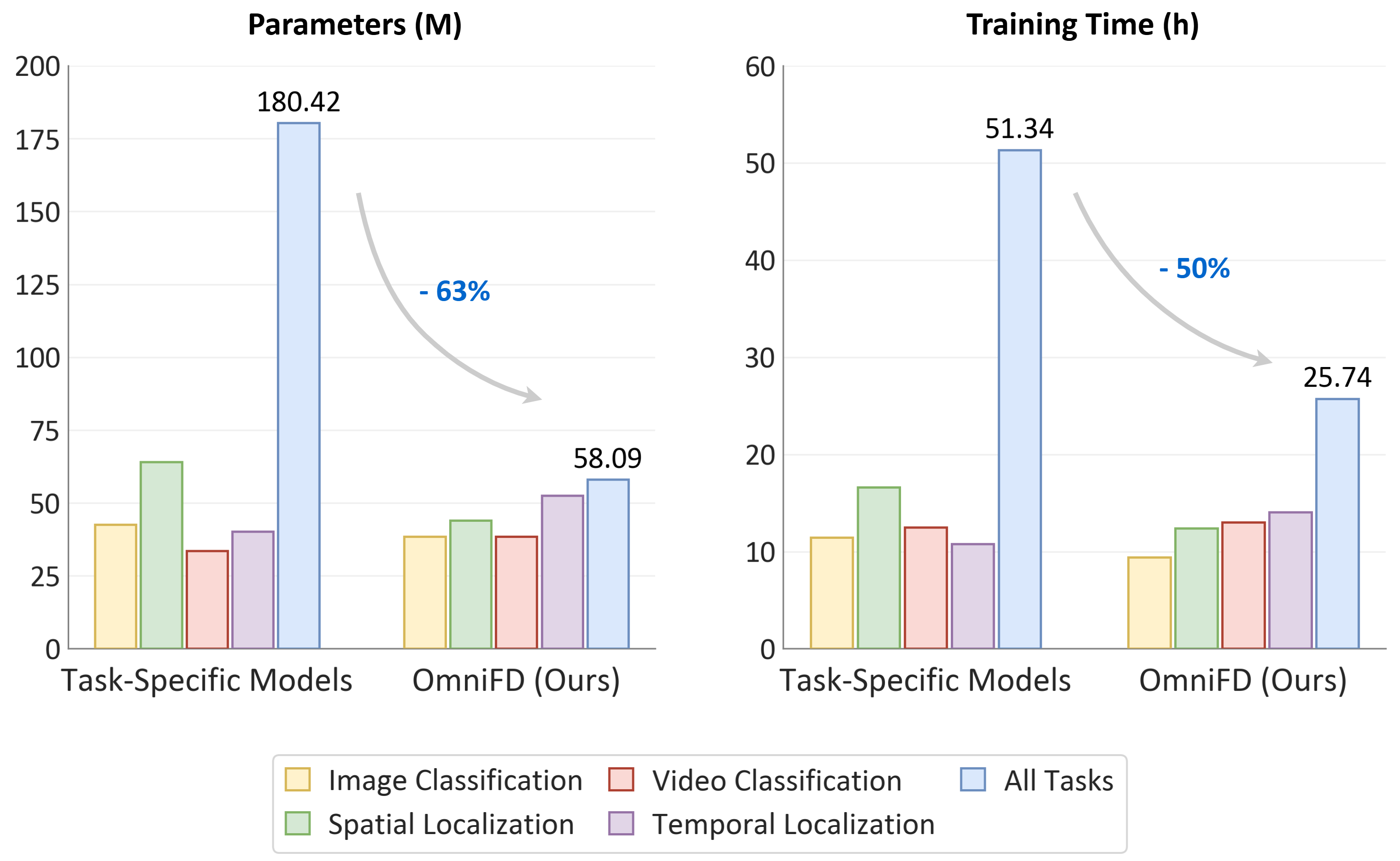}
        \end{minipage}}
    \caption{
    Advantages of the proposed unified model against task-specific models.
    (a) \textbf{Performance analysis.} 
    The evaluation covers representative benchmark datasets for each task, including \cite{forgerynet_he2021, faceforensics++_rossler2019, celeb_li2020celeb,deeperforensics_jiang2020,fmld_kong2022,wilddeepfake_zi2020,lavdf_cai2023}, and the latest state-of-the-art methods \cite{laanet_nguyen2024,ucf_yan2023,sparsevit_su2025,dadf_lai2023,mintime_coccomini2024,dfgaze_peng2024,ummaformer_zhang2023} are selected for comparison.
    The proposed OmniFD achieves superior performance within a single unified framework on all benchmarks of the four tasks.
    Detailed results are presented in Tables \ref{table:sota}, \ref{table:sota_image}-\ref{table:sota_temporal}.
    (b) \textbf{Efficiency analysis.} Here we compare representative task-specific models with OmniFD in terms of model parameters and training time for handling a single FFD task or all tasks jointly. OmniFD greatly reduces parameter redundancy and computational overhead compared with task-specific models. More detailed results are provided in Table~\ref{table:efficiency}. }
    \label{figure:overview}
\end{figure*}

A unified model \cite{omnivore_girdhar2022,tubevit_piergiovanni2023,ofa_wang2022,pix2seq_chen2022,imagebind_girdhar2023, allinone_wang2023, omnivid_wang2024} that incorporates multiple modalities and related tasks has attracted attention in visual understanding and natural language processing fields and demonstrated its advantages. 
The above-mentioned challenges in FFD could be addressed by developing a unified framework to process different tasks.
Fundamentally, all four FFD tasks (i.e., image classification, video classification, spatial localization, and temporal localization) aim to identify forgery artifacts, differing only in modalities and annotation granularity. 
Existing unified models \cite{allinone_wang2023, omnivid_wang2024} mainly concentrate on tasks within a single modality, or are designed as classification frameworks \cite{omnivore_girdhar2022, tubevit_piergiovanni2023, imagebind_girdhar2023} for multiple modalities in natural scenarios, making them unsuitable for unifying FFD tasks across images and videos.
There are preliminary studies \cite{m2tr_wang2022, forgerynet_he2021, uvif_liu2024} on unified FFD frameworks, but they address no more than two tasks and show limited generalization capability.

To address these limitations, we propose OmniFD, a unified framework for face forgery detection.
OmniFD is grounded in the principle of multi-task learning \cite{multitask_learning}, which encourages shared representation learning and jointly optimization of multiple objectives.
It aims to provide a unified and generalizable solution that enhances both performance and computational efficiency, as illustrated in Figure~\ref{figure:overview}.
The contributions of this paper include:%
\begin{itemize}
\item To our knowledge, it is the first attempt to address all four FFD tasks within a unified solution, eliminating the need for task-specific model designs and improving computational efficiency.
\item A unified encoder is designed to unify representation learning across images and videos by modeling spatial and temporal forgeries with shared parameters.
\item The framework leverages multi-task learning and cross-task interaction module to facilitate knowledge transfer among related tasks, allowing supervision from a task to reinforce another, such as between image and video classification.
\item Extensive experiments on multiple FFD datasets demonstrate that OmniFD achieves state-of-the-art performance with strong generalization across tasks. Its efficiency and scalability make it well-suited for real-world deployment.
\end{itemize}

\section{Related Work}

\subsection{Face Forgery Detection}

Face forgery detection \cite{faceforensics++_rossler2019,survey_ijcv_juefei2022} aims to identify manipulated or synthesized faces in images and videos, which is essential for mitigating misinformation and identity fraud risks.
Early studies \cite{mesonet_afchar2018, faceforensics++_rossler2019} focused on image classification, using convolutional neural networks (CNNs) \cite{xception_chollet2017, efficientnet_tan2019, resnest_zhang2022} to extract features from static images for binary classification. Some approaches leveraged local textures \cite{gramnet_liu2020,madd_zhao2021}, frequency domain analysis \cite{frequency_qian2020, frequency_li2021, f2trans_miao2023,ddl_sun2025}, and inconsistency information \cite{face_x-ray_li2020, ict_consistency_dong2022, self_blended_shiohara2022, recce_cao2022} to improve accuracy.
With the rise of deep generative models and face manipulation techniques, recent works \cite{ucf_yan2023, ual_wu2023generalizing, sfdg_wang2023dynamic, laanet_nguyen2024} emphasize improving generalization against unknown forgery attack types.
These image-based methods were also used to process video inputs by extracting frames from video clips and applying image classification models, while overlooking temporal dependencies present in videos. In contrast, other approaches \cite{ictu_li2018,lips_haliassos2021, biological_ciftci2020, spatiotemporal_gu2021,  uniformer_li2022, exploiting_wang2023, tall_xu2023, mintime_coccomini2024, dfgaze_peng2024, dfd_fcg_han2025} process entire video clips directly for video classification. Early methods employed recurrent neural networks (RNNs) to identify inconsistencies such as eye-blinking \cite{ictu_li2018}, lip motions \cite{lips_haliassos2021} and biological signals \cite{biological_ciftci2020}. Subsequent methods \cite{spatiotemporal_gu2021, exploiting_wang2023, tall_xu2023, mintime_coccomini2024, dfd_fcg_han2025} employed 3D CNNs and vision transformers to capture enhanced temporal dependencies for video forgery detection.

While image and video classification methods have shown promising performance and generalization capabilities on forgery benchmarks, they often lack interpretability for human users and fail to meet real-world application needs. 
To address this, researchers have explored fine-grained forgery detection tasks in spatial and temporal dimensions. Specifically, spatial localization \cite{faceforensics++_rossler2019,dffd_dang2020detection} predicts manipulated pixels in an image, while temporal localization \cite{forgerynet_he2021, lavdf_cai2023} detects forged segments in a video clip. These tasks require detailed annotations and improved representation, and specialized benchmarks \cite{lavdf_cai2023, fmld_kong2022, avdeepfake1m_cai2024} and methods \cite{fmld_kong2022, lavdf_cai2023, bsn_lin2018, tadtr_liu2022, tridet_shi2023, re2tal_zhao2023, adatad_liu2024, ba-tfd_cai2022, ummaformer_zhang2023, dadf_lai2023,upernet_xiao2018,hrnet_wang2020,iml_vit_ma2023,mesorch_zhu2025,sparsevit_su2025} are proposed. As some datasets \cite{faceforensics++_rossler2019, forgerynet_he2021} contain annotations of multiple forgery detection tasks, methods \cite{m2tr_wang2022, forgerynet_he2021, uvif_liu2024} integrate annotations of different tasks, e.g., image classification and spatial localization \cite{m2tr_wang2022}, for joint training.
However, they prioritize one task and use others as auxiliary supervision, which limits their generalization to other tasks.

To summarize, current forgery detection methods are usually designed and trained separately for each specific task, limiting cross-task or cross-modal model reuse. While all tasks focus on detecting forgery-related cues at different levels, their interrelations remain underexplored.

\subsection{Unified Models}
Unified models \cite{omnivore_girdhar2022,
tubevit_piergiovanni2023,
ofa_wang2022,
pix2seq_chen2022,
imagebind_girdhar2023, allinone_wang2023, omnivid_wang2024} have gained significant attention recently, standing out for their versatility and generalization capability. Unified vision models focus on processing different input modalities or vision tasks using a single model. Existing studies can be categorized into unified architectures and unified learning frameworks.

Unified architectures employ a single backbone to extract generic representations from various modalities, such as images and videos. As a video clip can be viewed as a sequence of image frames, a few methods \cite{tsm_lin2019, timesformer_bertasius2021, tubevit_piergiovanni2023} proposed to adapt 2D CNN or Transformer backbones from image tasks to process video data. OMNIVORE \cite{omnivore_girdhar2022} proposed a unified 4D representation for images, videos, and single-view 3D data, and utilized a Transformer model with shared parameters to process all modalities simultaneously. Besides, recent methods \cite{ imagebind_girdhar2023, allinone_wang2023} also proposed to align various data modalities, such as image, video, text, and audio, into a unified feature space via share Transformers. The unified architectures demonstrate promising performance and generalization capabilities, which also pave the way to process facial images and videos together in face forgery detection.

Unified learning frameworks utilize a single model or formulation to tackle a suite of related vision tasks. Most existing work focuses on tasks within the same modality. For example, object detection and segmentation methods \cite{maskrcnn_he2017, multi_task_loss_kendall2018} followed a multi-task learning paradigm \cite{multitask_learning} and jointly trained a shared backbone on a few prediction tasks. Later studies \cite{unit_hu2021, unified_retrieval_yan2023, sam_kirillov2023} first unified the annotation format of various tasks and then designed a shared decoder for making predictions. With the rapid progress in large language models (LLMs) \cite{gpt_radford2019, llama_touvron2023} and vision-language pretraining \cite{clip_radford2021, flamingo_alayrac2022, blip2_li2023}, recent methods \cite{pix2seq_chen2022,ofa_wang2022,  omnivid_wang2024} transform task predictions into text sequence predictions, thereby leveraging LLMs to achieve unified decoding across diverse visual tasks.
Nevertheless, the exploration of unified learning frameworks for cross-modal tasks, particularly involving both images and videos, remains limited. While some methods \cite{omni_duan2020, omnivore_girdhar2022, tubevit_piergiovanni2023} jointly process images and videos, they solely focus on classification. Consequently, unified models for face forgery detection across different tasks and modalities remain unexplored.

\section{Method}

\begin{figure*}[!t]
    \centering
    \includegraphics[width=0.95\linewidth]{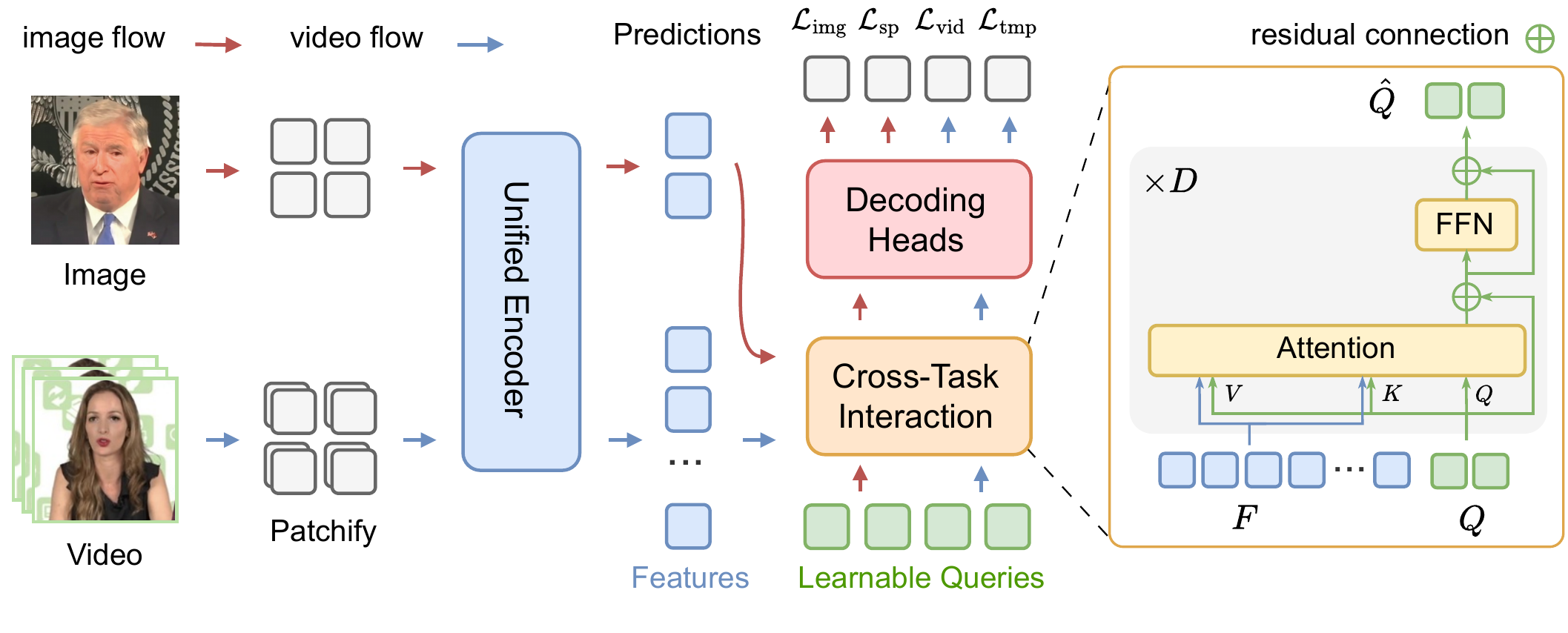}
    \caption{
    Overview of the proposed OmniFD framework. (1) The unified encoder processes both image and video inputs to extract shared spatiotemporal features. (2) The cross-task interaction module employs learnable queries that interact with the features to model inter-task dependencies. (3) The decoding heads transform the refined representations into predictions for four FFD tasks. The framework is optimized end-to-end under multi-task learning paradigm.
    }
    \label{figure:framework}
\end{figure*}

The proposed OmniFD framework provides a unified FFD solution across image and video modalities. It jointly addresses four FFD tasks within a single architecture. This formulation enables unified representation learning and promotes knowledge transfer among related tasks.
The framework consists of three components:
1) a unified encoder that extracts generalized spatiotemporal features from images and videos,
2) a cross-task interaction module that models task dependencies through transformer-based reasoning, and
3) task-oriented decoding heads that produce predictions for four FFD tasks.
The model is optimized end-to-end under a multi-task learning paradigm.
The remainder of this section is structured as follows: the unified representation across modalities is described in Section~\ref{section:encoder}, the cross-task interaction module in Section~\ref{section:interaction}, the task-oriented decoding in Section~\ref{section:head}, and the end-to-end optimization strategy in Section~\ref{section:optimization}.

\subsection{Unified Representation Across Modalities}
\label{section:encoder}

Our goal is to utilize a shared backbone for feature extraction across four FFD tasks through multi-task learning. Since all four tasks aim to identify forgery-related artifacts, using a shared backbone promotes unified representation learning and improves parameter efficiency. The main challenge arises from the heterogeneous input modalities, i.e., images and videos, and the different information priorities among the four tasks. Image-related tasks emphasize spatial cues, particularly spatial localization that relies on multi-scale features, whereas video-related tasks must also capture temporal information.

To address this challenge, we introduce a unified encoder that processes both image and video data to extract unified representations.
Inspired by \cite{omni_duan2020}, the input data is converted into a 4D tensor $\mathbf{X}\!\in\!\mathbb{R}^{T\times H\times W\times3}$ (with $T\!=\!1$ for images and $T\!>\!1$ for videos), divided into smaller patches and then fed into a Swin Transformer \cite{swin_liu2021}. The unified encoder leverages local self-attention together with spatial and temporal positional encodings to produce feature maps that capture fine-grained spatial information and temporal dynamics. Consequently, it generates spatiotemporal features \(\{\mathbf{F}^l\}_{l=1}^L\) from \(L\) backbone stages.
Since images and videos share the same parameters during feature extraction, the unified encoder learns generalized representations across both modalities.

Typical 2D CNN backbones can also be used to process image and video inputs with shared parameters, as described in \cite{tsm_lin2019, uvif_liu2024}, but they are less effective than the 4D tensor-based approach. We will make comparisons and discuss in the experiment sections.

\subsection{Cross-Task Interaction Module}
\label{section:interaction}
Multi-task learning \cite{multitask_learning} improves generalization by leveraging inductive bias that favors hypotheses useful across related tasks.
The inductive bias is manifested through auxiliary training signals, which guide the model to learn representations that capture relationships shared among tasks instead of optimizing each task independently.
This inductive bias is crucial in multi-task FFD, where different tasks rely on common manipulation artifacts.
The shared cues promote consistent representations across tasks, allowing knowledge from related tasks to reinforce the target task. 
For instance, detecting a forged frame in a video can aid in verifying the authenticity of the entire video clip.
The previous step employs a unified encoder to learn generalized representations, and the current step focuses on explicitly capturing interactions among FFD tasks.

Inspired by \cite{detr_carion2020,flamingo_alayrac2022}, we propose a transformer-based cross-task interaction module that employs learnable queries to model relationships among tasks, as illustrated in Figure \ref{figure:framework}.
Specifically, we introduce a set of learnable queries $\mathbf{Q}\in\mathbb{R}^{N \times C}$ that act as shared latent variables capturing dependencies among tasks, where $N$ and $C$ denote the number and dimension of queries, respectively. 
The queries interact with the feature maps $\{\mathbf{F}^l\}_{l=1}^L$ through attention mechanisms and are refined into $\hat{\mathbf{Q}}$ over $D$ transformer layers.
The refined queries are optimized to predict each FFD task, which drives them to encode forgery artifacts shared among FFD tasks.

The interaction among FFD tasks is achieved within the cross-task interaction module. The queries and feature maps are concatenated to form keys and values, allowing each query to aggregate information from others and enhance its own representation through the attention mechanism, as illustrated in Figure \ref{figure:framework}. The process is formulated as:
\begin{align}
\mathbf{K} &= \mathbf{V} = \text{Concat}\big (\mathbf{Q}, \{\mathbf{F}^l\}_{l=1}^L\big ),\\
\label{eq:interaction}
\mathbf{\hat Q} &= \text{LN} \big( \mathbf{Q} + \text{Attention}(\mathbf{Q}, \mathbf{K}, \mathbf{V}) \big),\\
\hat{\mathbf{Q}} &= \text{LN} \big( \mathbf{\hat Q} + \text{FFN}(\mathbf{\hat Q}) \big),
\end{align}
where $\mathbf{K}$, $\mathbf{V}$ are the key and value tensors, $\text{LN}$ denotes LayerNorm and FFN is Feedforward Network.
Through this process, the refined queries $\hat{\mathbf{Q}} \in \mathbb{R}^{N\times C}$ encode shared forgery patterns across FFD tasks.

Our cross-task interaction module employs a distinct mechanism compared to prior approaches \cite{detr_carion2020, perceiver_jaegle2021, flamingo_alayrac2022}. Existing methods either perform dimension reduction of input sequences \cite{flamingo_alayrac2022} or assign a single query to handle one specific prediction task \cite{detr_carion2020, perceiver_jaegle2021}, our cross-task interaction process allows learnable queries to adaptively model inter-task representations. These queries jointly address multiple FFD objectives through shared feature reasoning, overcoming the limitations of task-specific parameterization.

\subsection{Task-Oriented Decoding Heads}
\label{section:head}
The refined queries obtained from the cross-task interaction module provide more compact representations compared with the original features.
To make the representations discriminative, we utilize the refined queries to predict all four FFD tasks, allowing supervision signals from multiple objectives to refine their representations.
As the training progresses, the refined queries are optimized into embeddings that capture forgery-related artifacts shared across FFD tasks.

\begin{figure}[!t]
    \centering
    \includegraphics[width=\linewidth]{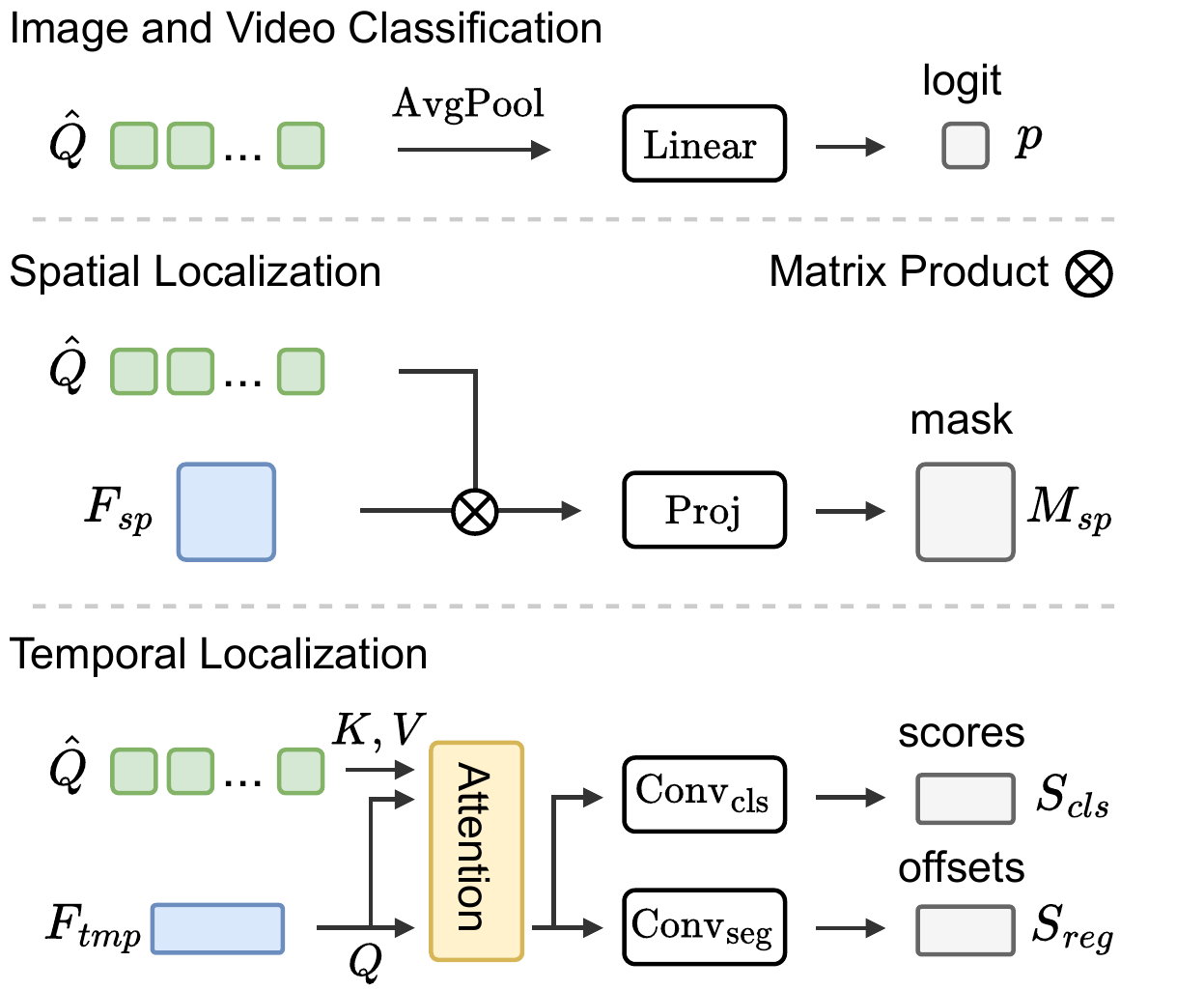}
    \caption{Illustration of decoding heads for different FFD tasks in OmniFD.}
    \label{figure:decoding}
\end{figure}

Based on this design, we employ four lightweight decoding heads that transform the refined queries into task-specific predictions.
In practice, separate decoding heads are used for each task to avoid unnecessary computational overhead when predictions for only one or several tasks are required.
As shown in Figure \ref{figure:decoding}, we adopts three decoding processes for the four FFD tasks.

\subsubsection{Image and Video Classification}
For image and video classification, average pooling \(\text{AvgPool}(\cdot)\) is applied to the refined queries $\hat{\mathbf{Q}} \in \mathbb{R}^{N\times C}$ to obtain a single feature vector, which is then passed through a linear layer \(\text{Linear}(\cdot)\):
\begin{equation}
    p = \text{Linear}(\text{AvgPool}(\hat{\mathbf{Q}})),
\end{equation}
where \(p \) represents the predicted logit for binary classification.

\subsubsection{Spatial Localization}
The spatial localization process first interpolates feature maps $\{\mathbf{F}^l\}_{l=1}^L$ to the same spatial resolution, concatenates them, and applies a $1\!\times\!1$ convolution layer to obtain the fused feature
$\mathbf{F}_{sp} \in \mathbb{R}^{HW\times C}$. The pixel-wise localization mask is then predicted as:
\begin{equation}
    \mathbf{M}_{sp} = \text{Proj}(\mathbf{F}_{sp}\hat{\mathbf{Q}}^\top),
\end{equation}
where \(\text{Proj}(\cdot)\) denotes a linear projection layer with weights $\mathbf{W}_{sp} \in \mathbb{R}^{N\times 1}$, and $\mathbf{M}_{sp} \in \mathbb{R}^{H \times W}$ 
represents the pixel-wise mask of forged region.

\subsubsection{Temporal Localization} 
Similar to typical temporal localization methods \cite{actionformer_zhang2022, tridet_shi2023}, the temporal feature $\mathbf{F}_{tmp} \in \mathbb{R}^{T \times C}$ is averaged from the final feature map $\mathbf{F}^{L}$.
It predict forged segments with two outputs: classification scores $\mathbf{S}_{cls} \in \mathbb{R}^{T \times 1}$ and boundary regression offsets $\mathbf{S}_{reg} \in \mathbb{R}^{T \times 2}$ (start and end time) at each temporal position $t \in \{ 1, ..., T \}$.

To incorporate the semantic information from the refined queries $\hat{\mathbf{Q}}$, transformer layer is applied to enhance the temporal feature $\mathbf{F}_{tmp}$ as:
\begin{align}
\mathbf{K} &= \mathbf{V} = \text{Concat}\big (\mathbf{F}_{tmp}, \hat{\mathbf{Q}}\big ),\\
\mathbf{\hat F}_{tmp} &= \text{LN} \big( \mathbf{F}_{tmp} + \text{Attention}(\mathbf{F}_{tmp}, \mathbf{K}, \mathbf{V}) \big), \\
\hat{\mathbf{F}}_{tmp} &= \text{LN} \big(\mathbf{\hat F}_{tmp}+\text{FFN}(\mathbf{\hat F}_{tmp}) \big).
\end{align}
This process resembles Eq. \ref{eq:interaction}, where $\mathbf{F}_{tmp}$ serves as the query.
It preserves the temporal dimension necessary for segment prediction at each temporal position. 
The prediction of forged segments is defined as:
\begin{align}
\mathbf{S}_{cls} &=\text{Conv}_\text{cls}(\hat{\mathbf{F}}_{tmp}), \\
\mathbf{S}_{reg} &= \text{Conv}_\text{reg}(\hat{\mathbf{F}}_{tmp}),
\end{align}
where $\text{Conv}_\text{cls}$ and $\text{Conv}_\text{reg}$ are 1D convolutional layers.

\subsection{End-to-End Optimization}
\label{section:optimization}
The OmniFD framework randomly samples annotated facial images and videos to form mini-batches and performs end-to-end optimization via multi-task learning. The overall loss $\mathcal{L}$ is defined as:
\begin{equation}
\begin{aligned}
\mathcal{L} =\ & \mathcal{L}_{\mathrm{img}} +  \mathcal{L}_{\mathrm{sp}} + \mathcal{L}_{\mathrm{vid}} + \mathcal{L}_{\mathrm{tmp}} ,
\end{aligned}   
\end{equation}
where $\mathcal{L}_{\mathrm{img}}$, $\mathcal{L}_{\mathrm{sp}}$, $\mathcal{L}_{\mathrm{vid}}$,  and $\mathcal{L}_{\mathrm{tmp}}$ represent losses for each
task, i.e., image classification, spatial localization, video classification, and temporal localization, respectively.

Specifically, $\mathcal{L}_{\mathrm{img}},\mathcal{L}_{\mathrm{vid}}$ employs standard cross entropy loss for binary classification, 
$\mathcal{L}_{\mathrm{sp}}$ computes the pixel-wise cross entropy loss, 
while $\mathcal{L}_{\mathrm{tmp}}$ combines focal loss for classification and DIoU loss for regression, as defined in \cite{actionformer_zhang2022}. Following common practice, $\mathcal{L}_{\mathrm{sp}}$ and $\mathcal{L}_{\mathrm{tmp}}$ are computed only from forged samples, while real samples are excluded from loss calculation.

Note that the unified design of OmniFD enables training with annotations from either a single task or any combination of tasks, without requiring both image and video data or full annotations for all four tasks. The experimental results on multiple FFD datasets validate this flexibility.

\section{Experiments}

\subsection{Experimental Setup}

\subsubsection{Datasets}
We evaluate our approach on five publicly available face forgery detection datasets.

\textit{ForgeryNet} \cite{forgerynet_he2021} is a large-scale and comprehensive dataset for face forgery detection, comprising over 2.9 million static images and 220k video clips. It includes 15 distinct image manipulation techniques and 8 distinct video manipulation techniques. The image and video subsets contain different subjects, forming two separate datasets.
ForgeryNet provides detailed annotations for all four FFD tasks: the image subset supports image classification and spatial localization, while the video subset supports video classification and temporal localization.
Following \cite{forgerynet_he2021}, 2.3 million images serve as the training set and 150k as the validation set, while 140k videos are used for training and 18k for validation.

\textit{FaceForensics++ (FF++) }\cite{faceforensics++_rossler2019} is a representative dataset for video classification, containing 1,000 real videos, each altered by four forgery methods. It is divided into training, validation, and test sets with 720, 140, and 140 videos, respectively. Since it provides annotations of manipulated regions and every frame in the fake videos is modified, FF++ is also adopted for image classification and spatial localization by extracting frames from the videos, as described in \cite{faceforensics++_rossler2019, fmld_kong2022}.

\textit{WildDeepfake (WDF)} \cite{wilddeepfake_zi2020} contains 7,314 deepfake video clips collected from the internet, covering diverse scenes, identities, and activities for video forgery classification. The dataset is split into 6,508 training and 806 testing samples \cite{wilddeepfake_zi2020}.

\textit{FMLD} \cite{fmld_kong2022} is a spatial forgery localization dataset comprising 120k synthetic and attribute-manipulated facial images generated using four GAN-based methods. The training, validation, and test sets follow an 8:1:1 split defined in \cite{fmld_kong2022}.

\textit{LAV-DF} \cite{lavdf_cai2023} is an audio-visual deepfake dataset containing manipulated segments in video or audio modalities. It includes 78k training videos and 26k testing videos. We follow \cite{lavdf_cai2023} and use the subset that excludes audio-only manipulation videos for evaluating temporal localization.

Seven additional datasets are used as test sets to evaluate the cross-domain generalization capability of image classification models, including \textit{CelebDF-v1 (CDFv1)} \cite{celeb_li2020celeb}, \textit{CelebDF-v2 (CDFv2)} \cite{celeb_li2020celeb}, \textit{DeeperForensics-1.0 (DF-1.0)} \cite{deeperforensics_jiang2020}, \textit{DeepFakeDetection (DFD)} \cite{faceforensics++_rossler2019}, \textit{DFDC} \cite{dfdc_dolhansky2020}, \textit{DFDCP} \cite{dfdc_dolhansky2020}, and \textit{FaceShifter (Fsh)} \cite{faceshifter_li2020advancing}. Following the protocol introduced in DeepfakeBench \cite{deepfakebench_yan2023}, we construct image classification datasets based on them by extracting facial frames from videos when necessary.

Experiments are conducted on these datasets. ForgeryNet dataset serves as the unified training source for all four FFD tasks. Its annotations across image and video modalities are used to evaluate OmniFD as a unified multi-task learning framework (Section \ref{section:sota}). FF++, WDF, LAV-DF, and FMLD datasets are employed to assess OmniFD’s generalization performance on individual tasks, whereas seven additional datasets (CDFv1, CDFv2, DF-1.0, DFD, DFDC, DFDCP, and Fsh) are used to evaluate its cross-domain generalization in image classification (Section \ref{section:generalization}).

\subsubsection{Evaluation Metrics}
We adopt commonly used evaluation metrics for each FFD task following established protocols.
Specifically, image classification is evaluated using Accuracy (Acc) and Area Under the ROC Curve (AUC) metrics, as in \cite{forgerynet_he2021, faceforensics++_rossler2019}.
For video classification, we report video-level Accuracy (Acc) and AUC following \cite{forgerynet_he2021}.
For spatial localization, two levels of Intersection over Union are used for ForgeryNet: IoU (threshold 0.1) and IoU$\mathrm{_{diff}}$ (threshold 0.01)\cite{forgerynet_he2021}, while Pixel-wise Binary Classification Accuracy (PBCA) and Inverse Intersection Non-Containment (IINC) are employed for other datasets \cite{faceforensics++_rossler2019, fmld_kong2022}, consistent with \cite{dffd_dang2020detection, fmld_kong2022}.
For temporal localization, Average Precision (AP), mean Average Precision (mAP) and Average Recall (AR) metrics are utilized, following the evaluation protocols in \cite{forgerynet_he2021, lavdf_cai2023}.

\subsubsection{Implementation Details}
For data processing, we follow \cite{faceforensics++_rossler2019, m2tr_wang2022, uvif_liu2024} and utilize RetinaFace \cite{retinaface_deng2020} to detect and extract facial regions from images and videos in ForgeryNet dataset.
The cropped faces are then resized to a resolution of \(\text{224} \times \text{224}\). 
For videos of ForgeryNet, we randomly sample 32 frames per video with a temporal stride of 4 during training.
Following \cite{faceforensics++_rossler2019, forgerynet_he2021, lavdf_cai2023}, we apply standard data augmentation techniques during training, including geometric transformations (random resizing, cropping, and horizontal flipping) and random perturbations (image compression and color distortion).

The proposed OmniFD is implemented with PyTorch and MMEngine \cite{mmengine2022}. We adopt the Swin Transformer \cite{swin_liu2021} as the unified encoder architecture, initialized with pretrained weights from ImageNet-1k \cite{imagenet_russakovsky2015} and Kinetics-400 \cite{kinetics_kay2017}, following \cite{omnivore_girdhar2022}. 
The cross-task interaction module uses 64 learnable queries ($N=\text{64}$) and 3 transformer layers ($D=\text{3}$).
During training, OmniFD jointly processes image and video data and constructs each mini-batch with 128 static images and 8 video clips from ForgeryNet \cite{forgerynet_he2021}.
Training is conducted on four NVIDIA Tesla V100 GPUs using the AdamW optimizer. The weight decay is set to 0.0001. The initial learning rate is set to 0.0001, followed by a cosine decay scheduler.
We train OmniFD for 50k iterations to ensure convergence.
The same training settings are used for all comparison methods unless specified otherwise.

\subsection{Comparison with State-of-the-Art}
\label{section:sota}

\begin{table*}[!t]
\centering
\caption{Comparison with the state-of-the-art methods on ForgeryNet \cite{forgerynet_he2021} video and image validation sets. 
Results marked with $^{\dag}$ are cited from \cite{forgerynet_he2021, uvif_liu2024, mintime_coccomini2024}.
Bold and underlined indicate the best and second-best results.
}
\label{table:sota}
\begin{tabular}{ll
*{4}{
>{\hspace*{0.25cm}\centering\arraybackslash}p{1.1cm} 
>{\centering\arraybackslash}p{1.1cm}<{\hspace*{0.25cm}}
}}
\toprule
\multirow{2}{*}[-0.75ex]{Method} & \multirow{2}{*}[-0.75ex]{Backbone} & \multicolumn{2}{c}{\textbf{Video Classification}} & \multicolumn{2}{c}{\textbf{Temporal Localization}} & \multicolumn{2}{c}{\textbf{Image Classification}} & \multicolumn{2}{c}{\textbf{Spatial Localization}} \\
\cmidrule(lr){3-4}\cmidrule(lr){5-6}\cmidrule(lr){7-8}\cmidrule(lr){9-10}
&  & Acc & AUC & AP@0.5 & mAP & Acc & AUC & IoU & IoU$_\mathrm{diff}$ \\
\midrule
SlowFast$^{\dag}$ \cite{slowfast_feichtenhofer2019} & SlowFast-50 & 88.78 & 93.88 & - & - & - & - & - & - \\
VideoSwin \cite{videoswin_liu2022} & Swin-T  & 84.22 & 91.73 & - & - & - & - & - & - \\
UniFormer \cite{uniformer_li2022} & UniFormer-S & 82.59 & 89.16 & - & - & - & - & - & - \\
TALL \cite{tall_xu2023} & Swin-B & 82.38 & 87.73 & - & - & - & - & - & - \\
UVIF$^{\dag}$ \cite{uvif_liu2024} & ResNet-101 & 86.57 & 94.42 & - & - & - & - & - & - \\
MINTIME-XC$^{\dag}$ \cite{mintime_coccomini2024} & Xception & 87.64 & 94.25 & - & - & - & - & - & - \\
DFD-FCG \cite{dfd_fcg_han2025} & ViT-L & 84.13 & 91.73 & - & - & - & - & - & - \\
\midrule
Xception$^{\dag}$ \cite{xception_chollet2017} & Xception & - & - & 68.29 & 62.83 & - & - & - & - \\
SlowFast+BSN$^{\dag}$ \cite{bsn_lin2018} & 3D ResNet-50 & - & - & 82.25 & \textbf{73.42} & - & - & - & - \\
ActionFormer \cite{actionformer_zhang2022} & SlowFast-101 & - & - & 74.73 & 55.81 & - & - & - & - \\
TriDet \cite{tridet_shi2023} & Swin-T & - & - & 74.08 & 59.69 & - & - & - & - \\
Re2TAL \cite{re2tal_zhao2023} & SlowFast-101 & - & - & 74.69 & 56.03 & - & - & - & - \\
UMMAFormer \cite{ummaformer_zhang2023} & ResNet-50 & - & - & 84.37 & 66.89 & - & - & - & - \\
AdaTAD \cite{adatad_liu2024} & VideoMAE-B & - & - & 78.06 & 60.96 & - & - & - & - \\
\midrule
EfficientNet$^{\dag}$ \cite{efficientnet_tan2019} & EfficientNet-B0 & - & - & - & - & 79.86 & 89.31 & - & - \\
Xception$^{\dag}$ \cite{xception_chollet2017} & Xception & - & - & - & - & 80.78 & 90.12 & - & - \\
ResNeSt$^{\dag}$ \cite{resnest_zhang2022} & ResNeSt-101 & - & - & - & - & \textbf{82.06} & \uline{91.02} & - & - \\
GramNet$^{\dag}$ \cite{gramnet_liu2020} & ResNet-18 & - & - & - & - & 80.89 & 90.20 & - & - \\
F3-Net$^{\dag}$ \cite{frequency_qian2020} & Xception & - & - & - & - & 80.86 & 90.15 & - & - \\
M2TR \cite{m2tr_wang2022} & EfficientNet-B4 & - & - & - & - & 81.41 & 90.67 & - & - \\
LAA-Net \cite{laanet_nguyen2024} & EfficientNet-B4 & - & - & - & - & 81.74 & 90.92 & - & - \\

\midrule
Xception+Reg$^{\dag}$ \cite{forgerynet_he2021} & Xception  & - & - & - & - & - & - & 89.55 & 67.57 \\
Xception+Unet$^{\dag}$ \cite{forgerynet_he2021} & Xception & - & - & - & - & - & - & 95.99 & 79.71 \\
HRNet$^{\dag}$ \cite{hrnet_wang2020} & HRNet-w48 & - & - & - & - & - & - & 96.27 & 88.73 \\
UperNet \cite{upernet_xiao2018} & HRNet-w48 & - & - & - & - & - & - & \uline{98.70} & 91.99 \\
IML-ViT \cite{iml_vit_ma2023} & ViT-B & - & - & - & - & - & - & 98.64 & 94.32 \\
Mesorch \cite{mesorch_zhu2025} & MiT-B3 & - & - & - & - & - & - & 98.49 & 93.96 \\
SparseViT \cite{sparsevit_su2025} & SparseViT & - & - & - & - & - & - & 98.62 & 94.40 \\
\midrule
multi-task (video) \cite{multitask_learning} & Swin-T  & 84.64 & 91.63 & 77.68 & 60.29 & - & - & - & - \\
multi-task (image) \cite{multitask_learning} & Swin-T  & - & - & - & - & 79.31 & 89.28 & 98.69 & \uline{94.80} \\
OMNIVORE-T \cite{omnivore_girdhar2022} & Swin-T & 87.43 & 94.98 & - & - & 78.45 & 88.60 & - & - \\
OMNIVORE-S \cite{omnivore_girdhar2022} & Swin-S & \uline{89.66} & \uline{96.31} & - & - & 80.83 & 90.33 & - & - \\
\midrule
OmniFD \textbf{(Ours)} & Swin-T & 88.75 & 95.44 & \uline{84.71} & 65.66 & 80.08 & 89.52 & 98.69 & 94.76 \\
OmniFD \textbf{(Ours)} & Swin-S & \textbf{90.10} & \textbf{96.57} & \textbf{87.32} & \uline{68.24} & \uline{81.89} & \textbf{91.10} & \textbf{98.72} & \textbf{94.93} \\
\bottomrule
\end{tabular}
\end{table*}

We compare the proposed OmniFD with state-of-the-art FFD methods on the ForgeryNet \cite{forgerynet_he2021} dataset, as shown in Table \ref{table:sota} and Figure \ref{figure:radar}. The comparison includes task-specific methods and multi-task learning methods across four FFD tasks, i.e., video classification, temporal localization, image classification, and spatial localization. OmniFD utilizes the same parameters for all tasks, except for the decoding heads.
Experimental results show that OmniFD achieves state-of-the-art performance across all four tasks.
Overall, OmniFD with Swin-S achieves the best performance on all four tasks (with the second-best mAP on temporal localization and accuracy on image classification), outperforming all single-task and dual-task baseline methods.

We further compare OmniFD with the conventional multi-task learning approach \cite{multitask_learning} and the unified OMNIVORE model \cite{omnivid_wang2024}. The multi-task learning approach uses annotations from the same modality (e.g., image classification and spatial localization), whereas OMNIVORE processes images and videos together for classification.
Specifically, the multi-task learning approach is implemented with a Swin-T backbone integrated with task-specific heads: linear layers for image and video classification, ActionFormer \cite{actionformer_zhang2022} for temporal localization, and FCN \cite{fcn_long2015} for spatial localization.
Compared with these methods, OmniFD achieves superior performance and supports multiple FFD tasks across different modalities in a single model.
These results indicate its effectiveness as a unified framework for face forgery detection.

\subsection{Advantage of Multi-Task Integration}
\label{section:multitask_integration}
OmniFD is flexible and versatile, which can be used for all four tasks, fewer tasks, or only one task to suit different application scenarios.
To explicitly demonstrate the advantages of fusing multiple FFD tasks, we test OmniFD in different training settings and compare the performance, i.e., OmniFD is trained on one task, a pair of tasks, and all four tasks. The experiment is conducted on the ForgeryNet dataset, and the results are shown in Table \ref{table:ablation_combinations}. The four FFD tasks, i.e., video classification, temporal localization, image classification, and spatial localization, are denoted as  \textit{Video}, \textit{Temporal}, \textit{Image}, and \textit{Spatial}, respectively.

From Table \ref{table:ablation_combinations}, three main findings can be concluded:

\textit{1) Cross-modal learning provides significant performance gains.}
Joint training of image and video classification (\textit{Image+Video}) leads to a clear improvement in video classification, achieving a +4.63\% gain in accuracy and +3.77\% in AUC over the video classification baseline (row 1).

\textit{2) Fine-grained temporal and spatial cues enhance higher-level forgery recognition.}
Specifically, joint training of video classification and temporal localization (\textit{Video+Temporal}) yields +0.95\% improvement in video classification accuracy, while the \textit{Image+Spatial} setting increases image classification accuracy by +0.40\%.

\textit{3) Full-task integration delivers the best overall improvement.}
OmniFD trained jointly on all four tasks (row 8) achieves the best overall performance, reaching 88.75\% accuracy in video classification and competitive results across other tasks.

The results demonstrate beneficial interactions among related and cross-modal tasks. Tasks with fine-grained annotations, such as temporal or spatial localization, provide detailed supervision that facilitates knowledge transfer and improves higher-level classification performance. In cross-modal learning, large-scale image data offer diverse visual priors that strengthen representation learning in the video domain, leading to more generalizable and transferable features. These benefits are further amplified when all tasks are trained jointly, allowing OmniFD to learn generalized representations across tasks and modalities. 

The spatial localization task shows limited improvement in the multi-task setting, primarily because other tasks lack supervision at a finer annotation level. Overall, OmniFD’s unified multi-task framework effectively exploits complementary information from FFD tasks, achieving robust and generalizable forgery detection across diverse scenarios.

\subsection{Efficiency Analysis}

\begin{table*}[!ht]
\centering
\caption{The effect of different task combinations in OmniFD on ForgeryNet \cite{forgerynet_he2021} validation set. 
}
\label{table:ablation_combinations}
\begin{tabular}{l*{4}{
>{\hspace*{0.3cm}\centering\arraybackslash}p{1.2cm} 
>{\centering\arraybackslash}p{1.2cm}<{\hspace*{0.3cm}}
}}
\toprule
\multirow{2}{*}[-0.75ex]{Setting} & \multicolumn{2}{c}{\textbf{Video Classification}} & \multicolumn{2}{c}{\textbf{Temporal Localization}} & \multicolumn{2}{c}{\textbf{Image Classification}} & \multicolumn{2}{c}{\textbf{Spatial Localization}} \\
\cmidrule(lr){2-3}\cmidrule(lr){4-5}\cmidrule(lr){6-7}\cmidrule(lr){8-9}
& Acc & AUC & AP@0.5 & mAP & Acc & AUC & IoU & IoU$_\mathrm{diff}$ \\
\midrule
\textit{Video} & 84.07 & 91.71 & - & - & - & - & - & - \\
\textit{Temporal} & - & - & 77.91 & 60.93 & - & - & - & - \\
\textit{Image} & - & - & - & - & 79.58 & 89.42 & - & - \\
\textit{Spatial} & - & - & - & - & - & - & \textbf{98.71} & \textbf{95.06} \\
\midrule
\textit{Video + Temporal} & 85.02 & 91.63 & \uline{79.06} & \uline{61.10} & - & - & - & - \\
\textit{Image + Spatial} & - & - & - & - & \uline{79.98} & \textbf{89.94} & \textbf{98.71} & \uline{94.85} \\
\textit{Video + Image} & \uline{88.70} & \textbf{95.48} & - & - & 79.91 & \uline{89.87} & - & - \\
\textit{All Tasks} & \textbf{88.75} & \uline{95.44} & \textbf{84.71} & \textbf{65.66} & \textbf{80.08} & 89.52 & \uline{98.69} & 94.76 \\
\bottomrule
\end{tabular}
\end{table*}

\begin{table}[!t]
  \centering
  \caption{Efficiency comparison of OmniFD and task-specific models when performing one to four FFD tasks. Results are based on a $224\times224$ image or 32-frame video on a V100 GPU.}
  \label{table:efficiency}
  \resizebox{1\linewidth}{!}{
  \begin{tabular}
    {clcccc}
    \toprule
    \makecell[c]{\rotatebox{90}{\textbf{}}}
      & Task
      & \makecell[c]{\textbf{Param}\\\textbf{(M)}}
      & \makecell[c]{\textbf{Flops}\\\textbf{(G)}}
      & \makecell[c]{\textbf{Latency}\\\textbf{(ms)}}
      & \makecell[c]{\textbf{Training}\\\textbf{Time (h)}} \\
    \midrule
    \multirow{7}{*}{\rotatebox[origin=c]{90}{Task-Specific}}
      & \textit{Image} (\textit{F3-Net}) & 42.52 & 10.95 & 9.51 & 11.45 \\
      & \textit{Spatial} (\textit{UperNet}) & 64.04 & 45.40 & 11.89 & 16.63 \\
      & \textit{Video} (\textit{SlowFast}) & 33.65 & 50.57 & 40.63 & 12.47 \\
      & \textit{Temporal} (\textit{UMMAFormer}) & 40.21 & 138.95 & 86.41 & 10.79 \\
      \cmidrule{2-6}
      & \textit{Image + Spatial} & 106.56 & 56.35 & 21.40 & 28.08 \\
      & \textit{Video + Temporal} & 73.86 & 189.53 & 127.04 & 23.26 \\
      & \cellcolor[HTML]{F0F0F0}\textit{All Tasks} & \cellcolor[HTML]{F0F0F0}180.42 & \cellcolor[HTML]{F0F0F0}245.87 & \cellcolor[HTML]{F0F0F0}148.44 & \cellcolor[HTML]{F0F0F0}51.34 \\
    \midrule
    \multirow{7}{*}{\rotatebox[origin=c]{90}{OmniFD (\textbf{Ours})}}
      & \textit{Image} & 38.42 & 7.21 & 11.86 & 9.39 \\
      & \textit{Spatial} & 44.01 & 11.51 & 13.47 & 12.39 \\
      & \textit{Video} & 38.42 & 121.84 & 52.11 & 13.02 \\
      & \textit{Temporal} & 52.50 & 122.71 & 62.43 & 14.06 \\
      \cmidrule{2-6}
      & \textit{Image + Spatial} & 44.01 & 11.51 & 13.47 & 12.80 \\
      & \textit{Video + Temporal} & 52.50 & 122.71 & 62.42 & 14.09 \\
      & \cellcolor[HTML]{F0F0F0}\textit{All Tasks} & \cellcolor[HTML]{F0F0F0}58.09 & \cellcolor[HTML]{F0F0F0}134.21 & \cellcolor[HTML]{F0F0F0}75.89 & \cellcolor[HTML]{F0F0F0}25.74 \\
    \bottomrule
  \end{tabular}
  }
  \ifjournal
  \else
  \vspace{-0.75eM}
  \fi
\end{table}

The efficiency of OmniFD stems from its unified architecture, which allows a single model to handle multiple FFD tasks during both training and inference.
In contrast, task-specific approaches use separate models for each task, leading to redundant parameters and extra computation when multiple tasks are involved.

Table \ref{table:efficiency} presents the quantitative results of efficiency analysis. We compare OmniFD (Swin-T) with four representative task-specific models, i.e., F3-Net \cite{frequency_qian2020}, UperNet \cite{upernet_xiao2018}, SlowFast \cite{slowfast_feichtenhofer2019}, and UMMAFormer \cite{ummaformer_zhang2023} (from Table \ref{table:sota}).
We report the number of parameters, FLOPs, inference latency, and training time for processing one to four FFD tasks. All metrics are measured using a 224$\times$224 image or a 32-frame video input on an NVIDIA V100 GPU.
The results indicate that OmniFD achieves comparable efficiency to task-specific models (rows 1–4) when trained on a single task, while its advantage becomes more pronounced as more tasks are involved.
Since task-specific models have distinct architectures, multiple forwardings are required to make predictions for different FFD tasks.
For instance, when processing all four tasks simultaneously (row 7), OmniFD reduces up to 63.40\% parameters, 45.41\% FLOPs, 48.87\% latency, and 49.86\% training time compared to task-specific models. 
Figure~\ref{figure:efficiency} further illustrates the efficiency improvements.
These results demonstrate OmniFD’s superiority in computational efficiency and its practicality for real-world deployment.

\begin{table*}[!t]
\centering
\caption{Comparison results and cross-dataset generalization on representative image classification benchmarks.}
\label{table:sota_image}
\begin{tabular}{lccccccccc}
\toprule
\multirow{2}{*}[-0.75ex]{Method} & \textbf{Intra} & \multicolumn{7}{c}{\textbf{Cross Dataset}} \\
\cmidrule(lr){2-2}\cmidrule(lr){3-10}
 & FF++ & CDFv1 & CDFv2 & DF-1.0 & DFD & DFDC & DFDCP & Fsh & Avg. \\
\midrule
Xception \cite{xception_chollet2017} & 96.37 & 77.94 & 73.65 & \uline{83.41} & \uline{81.63} & 70.77 & 73.74 & 62.49 & 74.80 \\
EfficientNet \cite{efficientnet_tan2019} & 95.67 & \uline{79.09} & 74.87 & 83.30 & 81.48 & 69.55 & 72.83 & 61.62 & 74.68 \\
F3-Net \cite{frequency_qian2020} & 95.92 & 70.93 & 67.86 & 55.31 & 76.55 & 63.26 & 69.42 & \uline{65.53} & 66.98 \\
RECCE \cite{recce_cao2022} & 96.21 & 76.77 & 73.19 & 79.85 & 81.19 & 71.33 & 74.19 & 60.95 & 73.92 \\
UCF \cite{ucf_yan2023} & \uline{97.05} & 77.93 & \uline{75.27} & 82.41 & 80.74 & \uline{71.91} & \uline{75.94} & 64.62 & \uline{75.55} \\
OmniFD (\textbf{Ours}) & \textbf{98.52} & \textbf{79.45} & \textbf{77.32} & \textbf{84.87} & \textbf{81.97} & \textbf{76.49} & \textbf{76.32} & \textbf{70.56} & \textbf{78.14} \\
\bottomrule
\end{tabular}
\end{table*}

\begin{table}[!t]
\centering
\caption{Comparison results of spatial localization on FF++ \cite{faceforensics++_rossler2019} and FMLD \cite{fmld_kong2022} testing sets.}
\label{table:sota_spatial}
\begin{tabular}{lcccc}
\toprule
\multirow{2}{*}[-0.75ex]{Method} & \multicolumn{2}{c}{FF++} & \multicolumn{2}{c}{FMLD} \vspace{0.1em} \\
\cmidrule(lr){2-3}\cmidrule(lr){4-5}
& PBCA ↑ & IINC ↓ & PBCA ↑ & IINC ↓ \\
\midrule
DFFD \cite{dffd_dang2020detection} & 94.85 & 4.57 & 98.72 & 3.31 \\
Locate \cite{fmld_kong2022} & 95.77 & 3.62 & 99.06 & \uline{2.53} \\
SAM \cite{sam_kirillov2023} & 92.97 & 4.25 & 97.29 & 3.40 \\
SAM+LoRA \cite{lora_hu2022} & 93.12 & 4.78 & 98.06 & 3.51 \\
DADF \cite{dadf_lai2023} & \uline{96.64} & \uline{3.21} & \uline{99.26} & 2.64 \\
OmniFD (\textbf{Ours}) & \textbf{98.60} & \textbf{1.91} & \textbf{99.46} & \textbf{1.55} \\
\bottomrule
\end{tabular}
\end{table}

\begin{table}[!t]
\centering
\caption{Comparison results of temporal localization on LAV-DF \cite{lavdf_cai2023} testing subset.}
\label{table:sota_temporal}
\resizebox{1\linewidth}{!}{
\begin{tabular}{
lccccc}
\toprule
Method & AP@0.5 & AP@0.75 & AP@0.95 & AR@10 & AR@20 \\
\midrule
BA-TFD \cite{ba-tfd_cai2022} & 85.20 & 47.06 & 0.29 & 59.32 & 61.19 \\
TadTR \cite{tadtr_liu2022} & 83.48 & 63.57 & 5.44 & 71.38 & 72.42 \\
TriDet \cite{tridet_shi2023} & 80.71 & 60.93 & 2.91 & 67.63 & 67.64 \\
BA-TFD+ \cite{lavdf_cai2023} & 96.82 & 86.47 & 3.90 & 79.15 & 79.60 \\
UMMAFormer \cite{ummaformer_zhang2023} & \uline{98.83} & \uline{95.95} & \uline{30.11} & \uline{92.32} & \uline{92.65} \\
OmniFD (\textbf{Ours}) & \textbf{98.96} & \textbf{96.81} & \textbf{59.69} & \textbf{97.08} & \textbf{97.47} \\
\bottomrule
\end{tabular}
}
\end{table}

\subsection{Generalization of the OmniFD Architecture}
\label{section:generalization}
The OmniFD is designed as a general-purpose architecture capable of generalizing across diverse FFD tasks and datasets. Beyond leveraging multi-task data for joint training, OmniFD can also address each task independently using the same architecture. To evaluate this generalization capability of OmniFD, we compare OmniFD with state-of-the-art methods across four FFD tasks on representative datasets beyond ForgeryNet, as summarized in Tables~\ref{table:sota_image}--\ref{table:sota_video} and Figure \ref{figure:radar}.
To ensure fair comparisons, we adopt identical evaluation protocols and use the same training data as other methods when training OmniFD on each task or dataset.

Table \ref{table:sota_image} presents the comparison results of image classification. We follow the same data pre-processing and cross-dataset evaluation protocol introduced in DeepfakeBench \cite{deepfakebench_yan2023},  train OmniFD (Swin-S) on the FF++ (c23) \cite{faceforensics++_rossler2019} dataset and evaluate the AUC performance across FF++ and other seven datasets: CDFv1 \cite{celeb_li2020celeb}, CDFv2 \cite{celeb_li2020celeb}, DF-1.0 \cite{deeperforensics_jiang2020}, DFD \cite{faceforensics++_rossler2019}, DFDC \cite{dfdc_dolhansky2020}, DFDCP \cite{dfdc_dolhansky2020}, and Fsh \cite{faceshifter_li2020advancing}. The results show that OmniFD achieves the best performance in both intra-dataset (98.52\% AUC) and cross-dataset (78.14\% average AUC) evaluation, demonstrating strong generalization capability in image forgery classification.
Similarly, Tables \ref{table:sota_spatial}, \ref{table:sota_temporal}, and \ref{table:sota_video} summarize the results on spatial localization, temporal localization, and video classification, respectively. Following the protocols in~\cite{dadf_lai2023,dfgaze_peng2024,lavdf_cai2023}, we evaluate spatial localization on FF++ and FMLD \cite{fmld_kong2022} datasets, video classification on FF++ and WDF \cite{wilddeepfake_zi2020} datasets, and temporal localization on LAV-DF \cite{lavdf_cai2023} dataset. OmniFD consistently achieves state-of-the-art performance across all FFD tasks over existing methods.

These results show that when OmniFD is trained on a single dataset for an individual task, it achieves state-of-the-art performance across benchmarks, demonstrating strong generalization capability and the simplicity of its unified architecture.

\subsection{Ablation Studies}
\subsubsection{Effectiveness of the Core Components}

We analyze the effectiveness of OmniFD's core components, i.e., unified encoder and cross-task interaction module, and the ablation results with Swin-T are presented in Table~\ref{table:ablation_components}.

The results show that both the unified encoder and cross-task interaction module are important for OmniFD’s unified modeling.
Compared to removing the unified encoder and training each task separately (row 1), using a unified encoder with shared weights for four tasks (row 3) brings significant performance gains, e.g., +4.72\% video classification accuracy and +4.73\% temporal localization mAP, demonstrating the benefit of unified representation learning with one shared encoder.
Meanwhile, row 2 removes the cross-task interaction module and replaces it with four vanilla task-specific decoders. Specifically, we use average pooling with linear layers for video and image classification, convolutional layers for spatial localization, and remove the transformer layer in temporal localization, making predictions of each task directly from the features without queries of the cross-task interaction module.
Compared to this setting, OmniFD with cross-task interaction module shows clear gains, e.g., +1.44\% mAP in temporal localization and +0.58\% accuracy in image classification, indicating that it improves knowledge transfer across tasks and promotes a unified decoding process.

\begin{table}[!t]
\centering
\caption{Comparison results of video classification on FF++ \cite{faceforensics++_rossler2019} and WDF \cite{wilddeepfake_zi2020} testing sets.}
\label{table:sota_video}
\begin{tabular}{lcccccc}
\toprule
\multirow{2}{*}[-0.75ex]{Method} & \multicolumn{2}{c}{FF++} & \multicolumn{2}{c}{WDF} \vspace{0.1em} \\
\cmidrule(lr){2-3}\cmidrule(lr){4-5}
 & Acc & AUC & Acc & AUC \\
\midrule
MADD \cite{madd_zhao2021} & 97.60 & 99.29 & 82.62 & 90.71 \\
RECCE \cite{recce_cao2022} & 97.06 & 99.32 & 83.25 & 92.02 \\
UAL \cite{ual_wu2023generalizing} & 98.13 & 99.56 & 84.16 & 92.56 \\
SFDG \cite{sfdg_wang2023dynamic} & 98.19 & 99.53 & 84.41 & 92.57 \\
DFGaze \cite{dfgaze_peng2024} & \textbf{98.80} & \textbf{99.91} & \uline{86.10} & \uline{93.00} \\
OmniFD (\textbf{Ours}) & \uline{98.57} & \uline{99.76} & \textbf{87.10} & \textbf{94.76} \\
\bottomrule
\end{tabular}
\end{table}

\begin{table}[!t]
\centering
\caption{Ablation study of the core components in the OmniFD framework.}
\label{table:ablation_components}
\begin{tabular}{cccccc}
\toprule
\multirow{2}{*}[-0.75ex]{\makecell{Unified \\ Encoder}} & \multirow{2}{*}[-0.75ex]{\makecell{Cross-Task \\ Interaction}} & \textbf{Video} & \textbf{Temporal} & \textbf{Image} & \textbf{Spatial} \\
\cmidrule{3-6}
 &  & Acc & mAP & Acc & IoU$_\mathrm{diff}$ \\
\midrule
- & \checkmark & 84.03 & 60.93 & 79.58 & \textbf{95.06} \\
\checkmark & - & 88.41 & 64.22 & 79.50 & 94.77 \\
\checkmark & \checkmark & \textbf{88.75} & \textbf{65.66} & \textbf{80.08} & 94.76 \\
\bottomrule
\end{tabular}
\end{table}

\subsubsection{Performance of Different Backbones}
We perform ablation experiments to evaluate OmniFD with different backbone architectures, as shown in Table \ref{table:backbone}. 
The backbones include representative CNN models, ResNet \cite{resnet_he2016} and ConvNeXt \cite{convnext_woo2023}, as well as the 4D tensor-based Swin Transformer \cite{swin_liu2021}.
We train four separate OmniFD models on each FFD task as baselines and compare them with the jointly trained model.

The main finding is that the jointly trained OmniFD consistently improves overall performance over the baselines across all backbones, significantly improving video classification accuracy by at least 4\%.
The backbones exhibit performance variations when trained separately for each task, mainly due to differences in architectural design.
Notably, the Swin Transformer, with its 4D tensor modeling and window-based attention, proves more effective for unified modeling in OmniFD than CNN-based backbones.
Among backbones of the same architecture (rows 5–8), the Swin-S variant achieves higher overall performance than Swin-T, indicating that larger-capacity models provide stronger generalization and enhance OmniFD’s representation learning across FFD tasks.
In summary, these results demonstrate the effectiveness of unified 4D tensor modeling and the scalability of OmniFD across backbone architectures.

\begin{table}[!t]
\centering
\caption{Ablation of different backbones for OmniFD on the ForgeryNet \cite{forgerynet_he2021} validation set. The subscript denotes improvement over the corresponding baseline.}
\label{table:backbone}
\resizebox{0.95\linewidth}{!}{

\begin{tabular}{lllll}
\toprule
\multirow{2}{*}[-0.75ex]{Method}
& \textbf{Video}
& \textbf{Temporal}
& \textbf{Image}
& \textbf{Spatial} \\
\cmidrule(l){2-5}
& Acc & mAP & Acc & IoU$_\mathrm{diff}$ \\
\midrule
ResNet-50 \cite{resnet_he2016} & 80.00 & 56.57 & 79.04 & 94.75 \\
\cellcolor[HTML]{F0F0F0}OmniFD (\textbf{Ours})
& \cellcolor[HTML]{F0F0F0}85.51$_{\uparrow5.5}$
& \cellcolor[HTML]{F0F0F0}57.57$_{\uparrow1.0}$
& \cellcolor[HTML]{F0F0F0}78.63$_{\downarrow0.4}$
& \cellcolor[HTML]{F0F0F0}94.39$_{\downarrow0.4}$ \\
\midrule
ConvNeXt-T \cite{convnext_woo2023} & 82.02 & 61.25 & 80.01 & 95.17 \\
\cellcolor[HTML]{F0F0F0}OmniFD (\textbf{Ours})
& \cellcolor[HTML]{F0F0F0}86.44$_{\uparrow4.4}$
& \cellcolor[HTML]{F0F0F0}64.37$_{\uparrow3.1}$
& \cellcolor[HTML]{F0F0F0}80.10$_{\uparrow0.1}$
& \cellcolor[HTML]{F0F0F0}94.87$_{\downarrow0.3}$ \\
\midrule
Swin-T \cite{swin_liu2021} & 84.03 & 60.93 & 79.58 & 95.06 \\
\cellcolor[HTML]{F0F0F0}OmniFD (\textbf{Ours})
& \cellcolor[HTML]{F0F0F0}88.75$_{\uparrow4.7}$
& \cellcolor[HTML]{F0F0F0}65.66$_{\uparrow4.7}$
& \cellcolor[HTML]{F0F0F0}80.08$_{\uparrow0.5}$
& \cellcolor[HTML]{F0F0F0}94.76$_{\downarrow0.3}$ \\
\midrule
Swin-S \cite{swin_liu2021} & 84.89 & 63.32 & 81.42 & 95.13 \\
\cellcolor[HTML]{F0F0F0}OmniFD (\textbf{Ours})
& \cellcolor[HTML]{F0F0F0}90.10$_{\uparrow5.2}$
& \cellcolor[HTML]{F0F0F0}68.24$_{\uparrow4.9}$
& \cellcolor[HTML]{F0F0F0}81.89$_{\uparrow0.5}$
& \cellcolor[HTML]{F0F0F0}94.93$_{\downarrow0.2}$ \\
\bottomrule
\end{tabular}

}

\end{table}

\subsection{Visualization}

\begin{figure}[!t]
\centering
\includegraphics[width=0.95\linewidth]{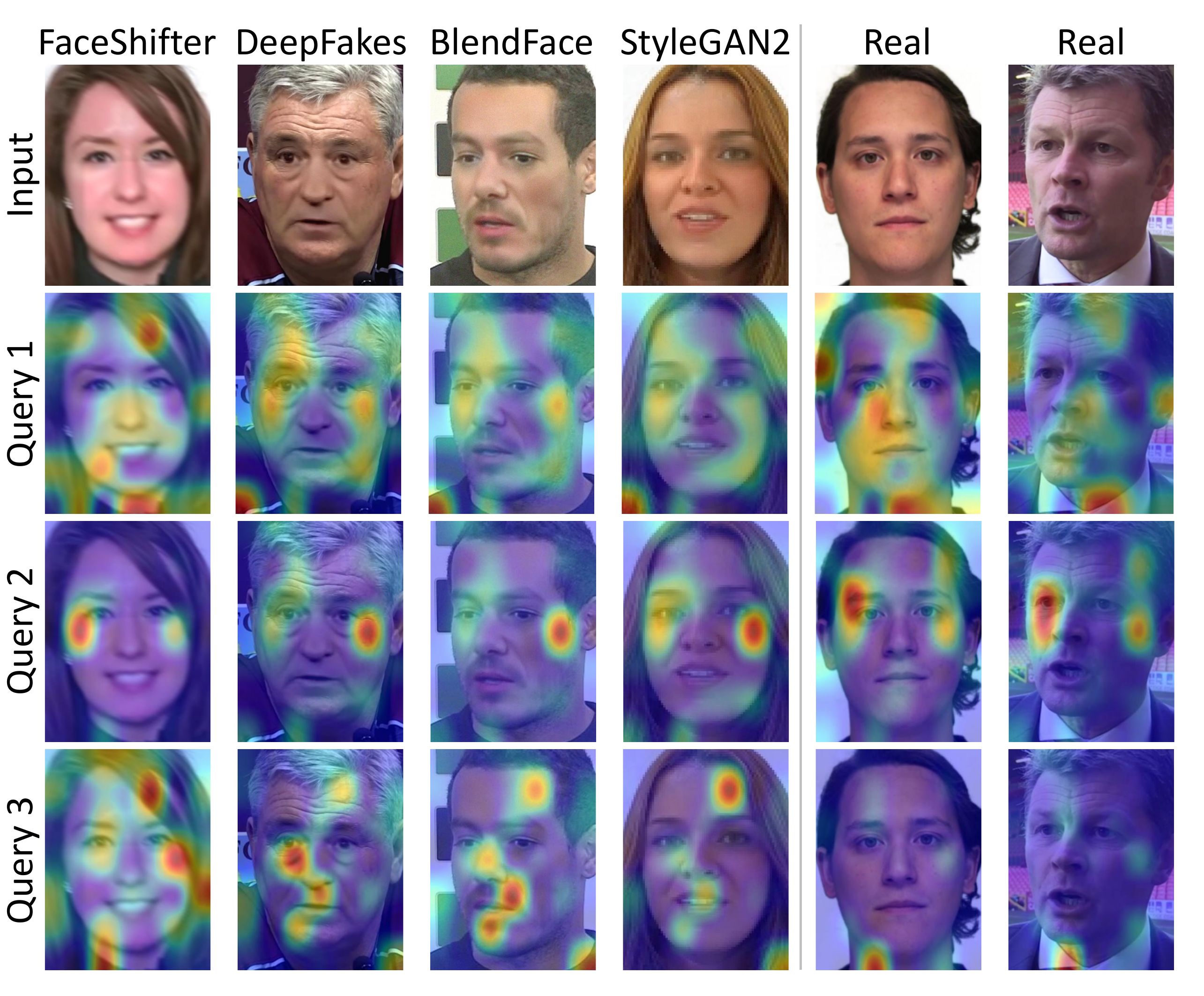}
\caption{
Visualization of attention maps between learnable queries $\mathbf{\hat Q}$ and features on forged and real faces. Samples of four types of forged faces and two real faces from ForgeryNet \cite{forgerynet_he2021} are included for comparison. The three bottom rows (Query 1, 2, and 3) show that different queries  learn to focus different facial attributes. Comparing the four left columns with the right two columns, it shows that the activation patterns of real faces differ from those of the forged faces.
}
\label{figure:attention}
\end{figure}

\subsubsection{Attention Map}
We visualize the attention maps between the learnable queries $\mathbf{Q}$ and input features $\mathbf{F}^L$ to analyze the mechanism of learnable queries in the cross-task interaction module, as shown in Figure~\ref{figure:attention}. The comparison includes four types of forged faces and two real faces from ForgeryNet~\cite{forgerynet_he2021}.
The visualization results show that different queries attend to distinct facial regions, from global facial contours to local attributes such as the eyes and mouth. For the same query, forged and real samples often exhibit different activation patterns, where certain queries (e.g., Query~3) provide strong cues for assessing facial authenticity.
These results indicate that the learnable queries capture forgery-related artifacts and encode them into generalized representations useful for face forgery detection.

\begin{figure}[!t]
\centering
\includegraphics[width=\linewidth]{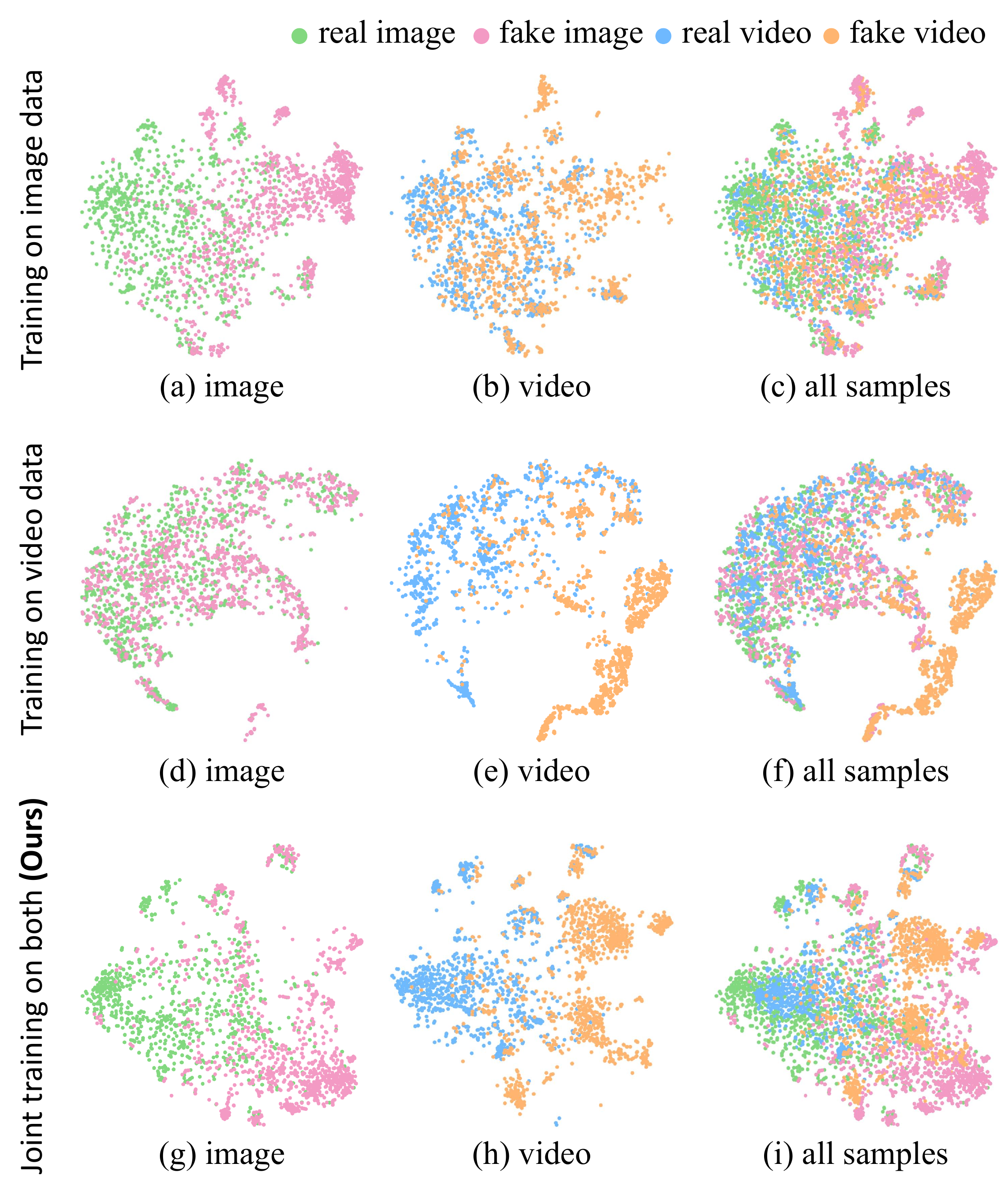}
\caption{T-SNE visualization of features in OmniFD under three settings: training on image data (a)-(c), training on video data (d)-(f), and joint training on both (g)-(i).
Each row corresponds to a different training setting, and the three columns show t-SNE embeddings for image, video, and all samples from the same model.
Single-modality training fails to achieve cross-modal generalization, while joint training yields compact clusters with consistent real–fake boundaries across image and video modalities.}
\label{figure:tsne}
\end{figure}

\subsubsection{T-SNE Visualization}

We further analyze the feature distribution and cross-modal alignment in OmniFD via t-SNE visualization, as shown in Figure~\ref{figure:tsne}.
We consider three training settings: training on image data, training on video data, and joint training on both.
We extract the t-SNE embeddings of the last-stage features $\mathbf{F}^{L}$ using the same model on both image and video samples from ForgeryNet~\cite{forgerynet_he2021}.

As shown in Figure~\ref{figure:tsne}(a)–(c), when trained only on image data, the model separates real and fake images, but video samples remain entangled. A similar pattern appears in Figure~\ref{figure:tsne}(d)–(f), where real and fake videos are separated while image samples are intermixed.
In contrast, jointly training OmniFD on image and video data, i.e., Figure~\ref{figure:tsne}(g)–(i), produces compact and well-aligned clusters for both modalities. Figure~\ref{figure:tsne}(i) shows clearer boundaries between real and fake samples across modalities, whereas those in Figure~\ref{figure:tsne}(c) and (f) are vague.
These results demonstrate that single-modality training fails to learn representations that generalize across modalities, making joint supervision essential for learning modality-invariant representations and improving cross-modal alignment.

\section{Conclusion}

We present OmniFD, the first unified model that simultaneously tackles four core face forgery detection tasks: image classification, video classification, spatial localization, and temporal localization. OmniFD eliminates the need for task-specific models by employing a unified encoder to extract generalized representations across different modalities and tasks. With the proposed cross-task interaction module and multi-task optimization, OmniFD effectively exploits inter-task correlations, achieving state-of-the-art performance on multiple benchmarks while demonstrating its efficiency and scalability. This provides a unified framework for practical and generalizable face forgery detection.
For future work, we will develop foundational models for face forgery detection from a data scaling perspective. Additionally, we will explore how to integrate multimodal data such as audio and text for more comprehensive face forgery analysis.

\bibliographystyle{IEEEtran}
\bibliography{main}

 




\vfill

\end{document}